%% file: egpaper.tex
\documentclass[10pt,twocolumn,letterpaper]{article}

\usepackage{wacv}
\usepackage{times}
\usepackage{epsfig}
\usepackage{graphicx}
\usepackage{amsmath}
\usepackage{amssymb}
\usepackage[skip=0.2\baselineskip]{caption}
\usepackage[accsupp]{axessibility}  % Improves PDF readability for those with disabilities.

\renewcommand\and{\kern15pt}

\usepackage{multirow,tabularx}

\def\wacvPaperID{1368} % Enter the WACV Paper ID here

\wacvfinalcopy % *** Uncomment this line for the final submission

\ifwacvfinal
\def\assignedStartPage{4012}% *** Enter the assigned starting page number (instead of 9876)
\fi

\ifwacvfinal
\usepackage[breaklinks=true,bookmarks=false]{hyperref}
\else
\usepackage[pagebackref=true,breaklinks=true,colorlinks,bookmarks=false]{hyperref}
\fi

\ifwacvfinal
\else
\fi

\begin{document}

%%%%%%%%% TITLE
\title{SpectraNet: Learned Recognition of Artificial Satellites from High Contrast Spectroscopic Imagery}

\author{
J. Zachary Gazak$^1$\thanks{jonathan.gazak.1.ctr@us.af.mil}
% {\tt\small jonathan.gazak.1.ctr@us.af.mil}

\and
Ian McQuaid$^2$
% \\ {\tt\small ian.mcquaid@centauricorp.com}

\and
Ryan Swindle$^1$
% \\ {\tt\small thomas.swindle.2.ctr@us.af.mil}

\and
Matthew Phelps$^1$
% \\ {\tt\small matthew.phelps.15.ctr@us.af.mil}

\and
Justin Fletcher$^1$\thanks{justin.fletcher.14.ctr@us.af.mil}
% \\ {\tt\small justin.fletcher.14.ctr@us.af.mil}

\\ $^{1}$United States Space Force Space Systems Command
\\ $^{2}$Air Force Research Laboratory}

\maketitle
%\thispagestyle{empty}

%%%%%%%%% ABSTRACT
\begin{abstract}
Effective space domain awareness requires positive identification of artificial satellites.  Current methods for extracting object identification from observed data require spatially resolved imagery which limits identification to objects in low earth orbits.  Many artificial Earth satellites, however, operate in geostationary orbits at distances which prohibit ground based observatories from resolving spatial information. This paper demonstrates an object identification solution leveraging modified residual convolutional neural networks to map distance-invariant spectroscopic data to object identity.  We report classification accuracies exceeding 80\% for a simulated 64-class satellite problem$-$even in the case of satellites undergoing constant, random re-orientation.  An astronomical observing campaign driven by these results returned accuracies of $\sim$72\% for a nine-class problem with an average of 100 examples per class, performing as expected from simulation.  We demonstrate the application of variational Bayesian inference by dropout, stochastic weight averaging (SWA), and SWA-focused deep ensembling to measure classification uncertainties$-$critical components in space domain awareness where routine decisions risk expensive space assets and carry geopolitical consequences.
\end{abstract}

%%%%%%%%% BODY TEXT
\section{Introduction}
\label{sec:intro}

The ideal data type for identifying resident space objects (RSOs; artificial satellites) is resolved imagery.  Analysts can easily interpret the information content of images and recent work has demonstrated that deep neural networks provide efficiency enhancements to signal extraction \cite{Kyono2020TowardsNetworks, Lucas2020AutomatedLearning,  Werth2020Quality-WeightedQWID}.  In addition, data collection is passive -- RSOs always reflect incident sunlight, and ground based telescopes collect without interfering with the target.

\begin{figure}[!h]
	\centering
	\includegraphics[width=0.155\textwidth]{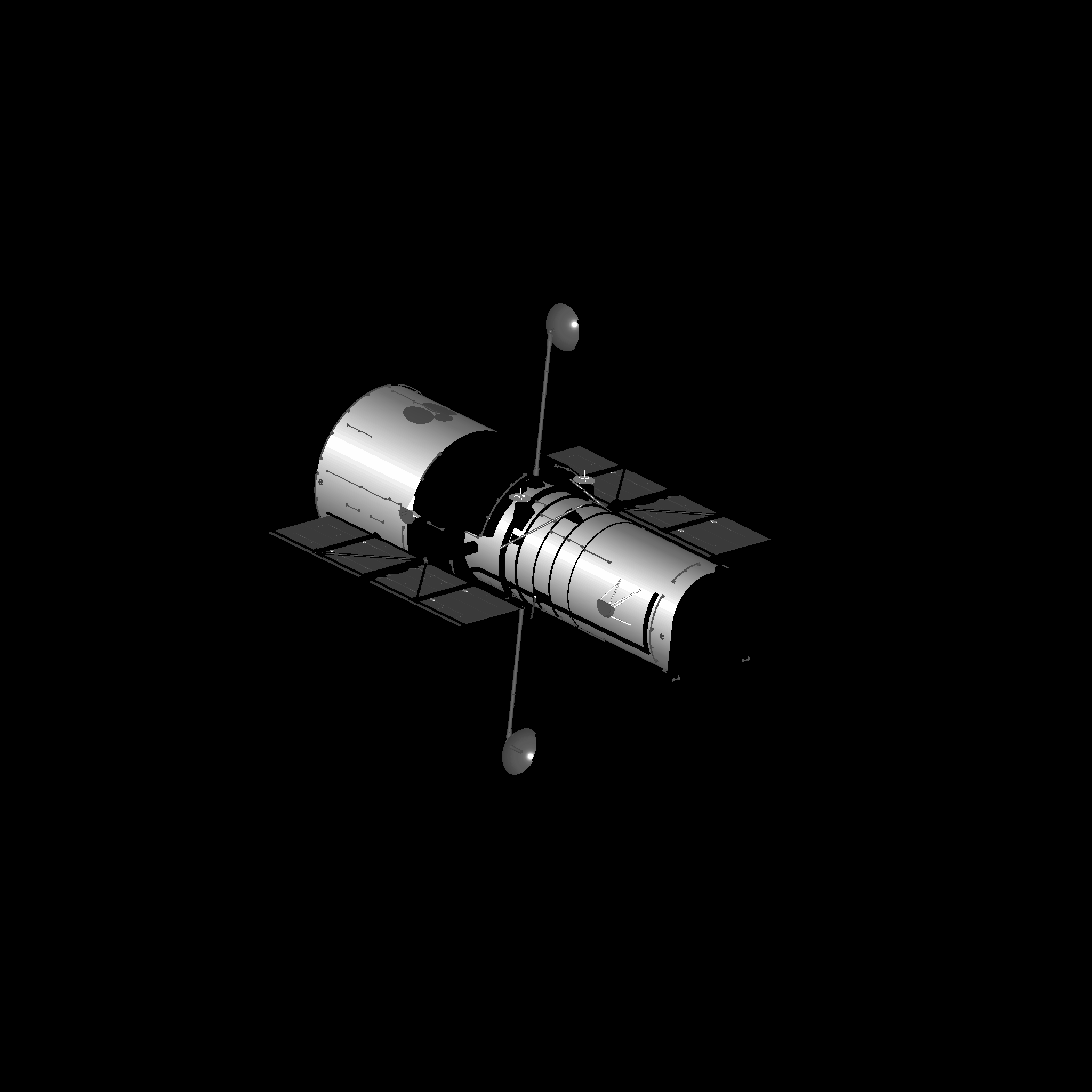}
	\includegraphics[width=0.155\textwidth]{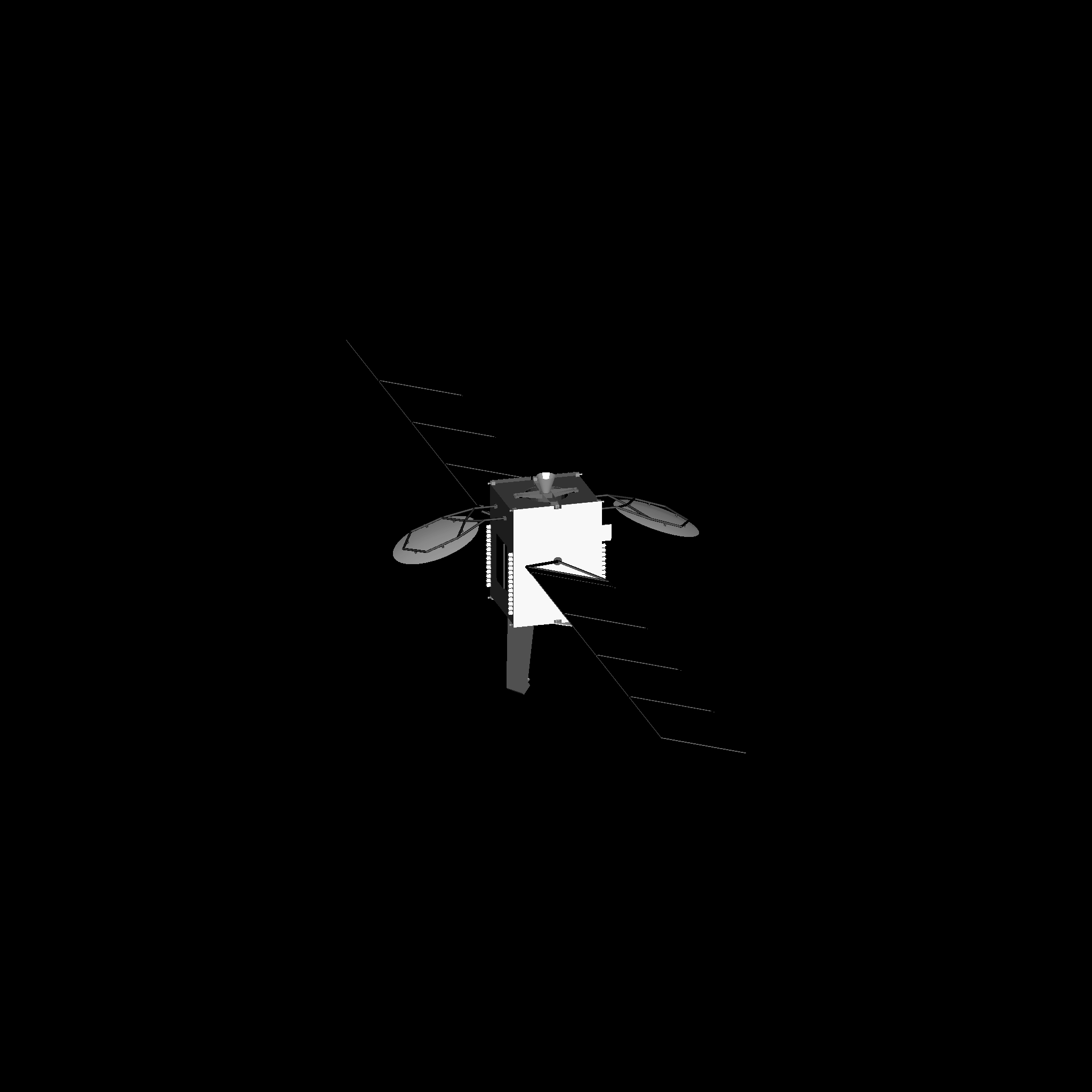}
	\includegraphics[width=0.155\textwidth]{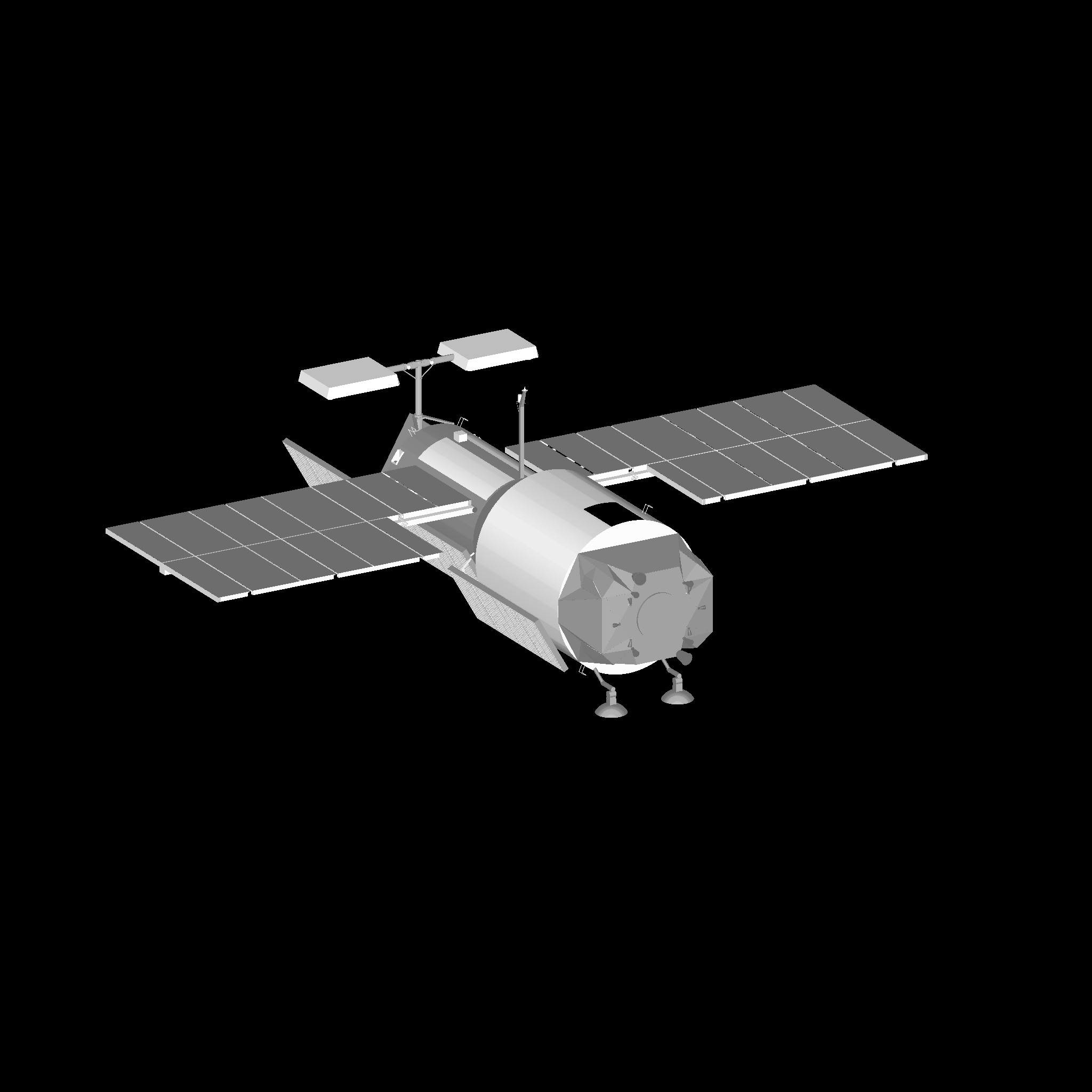}
	\caption{Pristine renders of three satellites, Hubble, DIRECTV, and Almaz \cite{Phelps2021InferringNetworks}.  Satellite spectral energy distributions are complex and vary strongly with orientation and illumination angle.}
    \label{fig:exampleimg}
\end{figure}

Unfortunately, smaller and more distant RSOs require increasingly large telescopes to resolve, placing the majority of RSOs beyond the limits of current imaging technology.  For positive identification of objects beyond low earth orbit (LEO; altitude $<$ 1000 km), new approaches are as necessary as they are elusive.  

\begin{figure}[!h]
	\centering
	\includegraphics[width=0.48\textwidth]{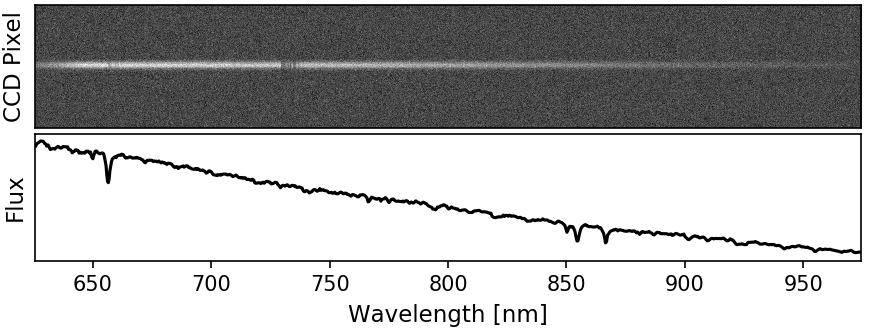}
	\caption{\textbf{Top panel}: A simulated raw FPA observation of 18 Scorpii, a star which is a close analog of our Sun \cite{CayreldeStrobel1996StarsSun}, such that this spectrum and FPA frame are typical of resident space objects reflecting solar radiation.  These raw frames are used to train models in this paper.  \textbf{Bottom}: The 1-D reduced spectrum of 18 Scorpii after a raw FPA frame is fully calibrated \cite{Bohlin2014DissertationMid-Infrared}.}
	\label{fig:examplespec} % Labels have to go after captions in h! or weird things can happen.-jrf
\end{figure}

One promising technique, spectroscopy, is both passive and distance-invariant, having been used to study the most distant objects in the universe for well over 50 years \cite{Kellermann2014THEAFTERMATH}.  In the field of space domain awareness, spectroscopy has lagged behind its potential for three reasons.  First, the inaccuracy of material reflection models (bi-directional reflection functions, or BRDFs) does not allow attribution of spectroscopic features to satellite materials and geometries.  Second, spectroscopic data is not easy to interpret (see Figure~\ref{fig:examplespec})$-$out of reach of human analysts even after extensive expert data reduction.  Third, underlying truth data on material and geometry are difficult to obtain as spectra are highly variable based on orientation (Figure~\ref{fig:exampleimg}).  These hurdles preclude transition of spectroscopic solutions beyond laboratory settings.

In this work we demonstrate \begin{it}learned\end{it} spectroscopic positive identification by modeling a high yield, low cost sensor and training convolutional neural networks (CNNs) on simulated output.  By eliminating reliance on both physics-based priors and exquisite data reduction, our technique (SpectraNet) identifies RSOs with accuracy exceeding 80\% under the most difficult condition of random axis orientation and for large numbers of satellite classes.

We verify simulated results by collecting on sky spectral observations of RSOs in geostationary (GEO; altitude $>$ 35000 km) orbits.  SpectraNet learns to classify objects in our on sky dataset with accuracy of $\sim$72\%, in agreement with simulated results given the limited ($\sim$100 examples per class) dataset.  In this paper we contribute:

\begin{itemize}
    \item A novel method capable of identifying spatially unresolved artificial satellites$-$a critical technology for space domain awareness$-$by allowing deep Bayesian residual networks to learn spectroscopic features from raw scientific imagery. These models can produce well calibrated softmax probabilities, enabling practical applications of SpectraNet at low sample counts.
    \item Both real and synthetic datasets representing the spectroscopic satellite identification application domain and baseline analysis of classifier performance on those datasets as a function of observation count (dataset size).\footnote{unifieddatalibrary.com/sfm/?expandedFolder=/SupportingData/MISS}
    \item A framework for designing spectroscopic positive identification systems by experimenting across number of classes, observation quality (signal to noise), and number of examples needed to achieve suitable classification performance.
\end{itemize}

In \S\ref{sec:related} we discuss topics related to this work, followed by our specific problem formulation in \S\ref{sec:formulation}.  We describe the datasets used for training in \S\ref{sec:datasets}, and the experiments conducted in \S\ref{sec:experiments}, before summarizing results in \S\ref{sec:conclusion}.

\section{Related Works}
\label{sec:related}

\subsection{Deep Learning for Space Domain Awareness}
\label{ssec:posid}

To our knowledge no practical methods have demonstrated positive identification of spatially unresolved RSOs, and thus any identification of objects in geostationary orbits are, at best, random. 

More generally, convolutional neural networks have been applied to object detection, detection of closely spaced objects, pose estimation, reconstruction of high resolution imagery, and segmentation of satellites \cite{Fletcher2019Feature-BasedNetworks, Gazak2020ExploitingNeural,  Okkelberg2021Self-SupervisedOrientation, Werth2020SILO:Satellites, Yang2021SemanticNetworks}. In these examples, solutions learned from high contrast scientific imagery solve problems faster and more effectively than physics based methods.

\subsection{Residual Networks}
\label{ssec:resnets}

The batch normalization and skip connections of Residual Networks (ResNets) have long been applied to increase the training speed and stability of deep neural networks \cite{He2015DeepRecognition}.  We adopt the ResNet-152 backbone from that work with adjustments (see \S\ref{sec:formulation}) to the initial convolutional layer to account for our non-standard input data shape described in \S\ref{sec:datasets}.  

\subsection{Bayesian Neural Networks}
\label{ssec:bayesnets}

Advantages of neural networks formulated to deliver probabilistic inference are tantalizing in problems with limited data.  Such Bayesian formulations, first introduced in 1992 by \cite{MacKay1992}, have become an active area of research as teams attempt to overcome difficult problems such as prior choice \cite{Wu2019DeterministicNetworks}, with evidence supporting both the need for descriptive priors in shallow networks \cite{Foong2020OnNetworks} and suggesting such computationally restrictive prior choices can be substituted for simple approximations when working with larger backbones \cite{Farquhar2020LibertyApproximations}.

Deep networks trained on small datasets are over specified, such that many parameter combinations provide high performance.  In contrast to classic point estimate deep networks$-$which \textbf{optimize} to a static parameter set$-$probabilistic Bayesian formulations rely on \textbf{marginalization}; the combination of many parameter settings scaled by posterior likelihoods.  We roughly categorize these techniques into single- and multi-basin ensembles to distinguish those which marginalize within a basin of attraction of one model, and those which marginalize over multiple trained instances of a model.

The simplest implementations introduce dropout after every network layer during training and inference \cite{Gal2016DropoutGhahramani}.  Such implementations are computationally efficient and mathematically equivalent to placing Bernoulli priors over every convolutional weight \cite{Gal2016BayesianInference}.  By decreasing active neurons, though, the capacity of a model is decreased.  We adopt this single-basin technique for large 64 class datasets to capitalize on training efficiency.

For smaller nine class datasets we adopt the perspective of \cite{Wilson2020BayesianGeneralization} that deep ensembles represent Bayesian marginalization.  In a single basin of attraction, \cite{Izmailov2018AveragingGeneralization} find that Stochastic Weight Averaging (SWA) shifts a deep network away from the highest performing training loss towards the center of the wider loss valley.  That center, offset from optimal training loss, should generalize better due to steep increases in loss near basin edges in stochastic gradient descent \cite{Izmailov2018AveragingGeneralization}.  Unlike SWA, in which a running mean of small batch model parameters is stored for inference, SWA-Gaussian (SWAG) samples multiple solutions from a SWA model by perturbing the SWA weights using a low rank and diagonal covariance approximation accumulated during training \cite{Maddox2019ALearning}.  SWAG defines a Gaussian posterior over network weights, while SWA implies that posterior by training with constant learning rate.

\begin{figure}[]
	\centering
	\includegraphics[width=0.49\textwidth]{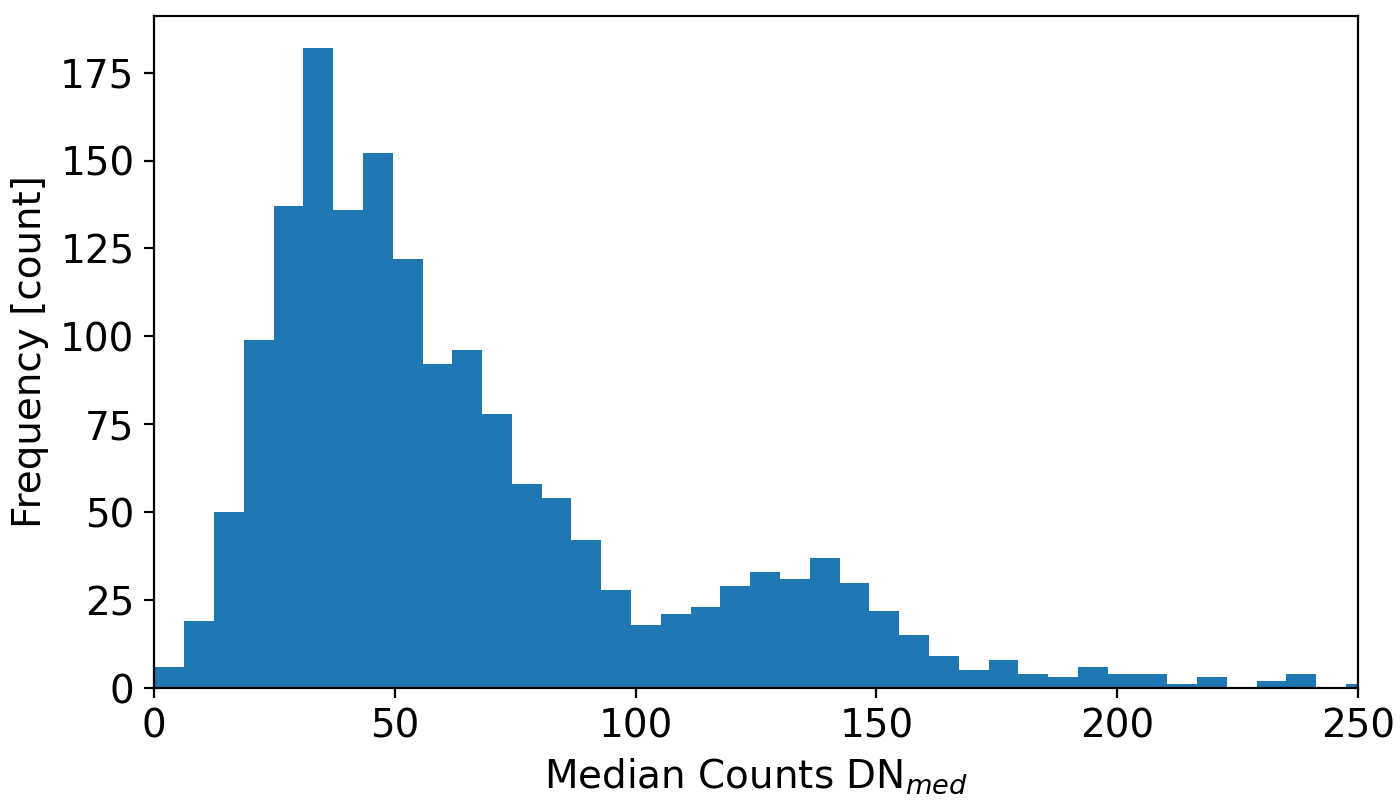}
	\caption{Distribution of signal to noise proxy DN$_{med}$ for the observed dataset in this work.  While lower values offer enticingly short exposure times and higher yield observations, this work demonstrates that classifier performance suffers dramatically below DN$_{med}$ of 50 (\S\ref{sec:experiments_sky}).}
	\label{fig:snr}
\end{figure} 

Finally, we extend SWA and SWAG by investigating multi-basin techniques multi-SWA and multi-SWAG, in which multiple training iterates are ensembled \cite{Wilson2020BayesianGeneralization}.  Ensembling across multiple models drives increased accuracy and calibration by sampling not only within basins of attraction but also across many unique basins.  We note that SWAG and multi-basin ensembling techniques require significantly more training and inference time (or computing power), and discuss these trade offs in \S\ref{sec:conclusion}.

\subsection{Model Calibration}

Deep networks are known to be poorly calibrated and highly overestimate classification confidence.  With limited data and the significant support of deep neural networks, many parameterizations provide functionally unique inference with similar performance; marginalizing over such diversity provides increased accuracy and better calibration \cite{Wilson2020BayesianGeneralization}.  Calibration can be further improved by tempering a Bayesian posterior (see Equations 2-3 of \cite{Wilson2020BayesianGeneralization}).  In practice this is achieved by raising output logits to the power T$^{-1}$ before producing sharper (T$<$1) or more diffuse (T$>$1) softmax probabilities.

We adopt expected calibration error (ECE) to measure the base and tempered calibrations of our ensembling studies \cite{Ovadia2019CanShift}. ECE is calculated by binning validation inferences by softmax probability of the predicted class and summing the weighted difference between the binned softmax prediction and the percentage of successful inferences in that bin.  That difference, ECE, becomes negligible ($\sim$0), when softmax probabilities are properly calibrated to validation predictions.

During experiment evaluation, we vary temperature 0.05 $\le$ T $\le$ 10 and report the best expected calibration error and temperature required to reach it.  T $>$ 1 represents overconfident trained models, while T $<$ 1 suggests underconfidence.

\begin{table}
\centering
\begin{tabular}{|l|rrr|}
\hline
Target &       DN$_{med} >$ 0 &  $>$ 50 & $>$ 100 \\
\hline
 GEO 1 &       274 &          235 &     142 \\
 GEO 2 &       127 &           54 &      14 \\
 GEO 3 &       219 &          119 &      54 \\
 GEO 4 &       357 &          128 &      32 \\
 GEO 5 &       128 &           67 &      16 \\
 GEO 6 &       215 &           85 &      15 \\
 GEO 7 &        93 &           90 &      88 \\
 GEO 8 &       320 &          161 &      21 \\
 \hline \hline
 Total &      1733 &          939 &     382 \\
\hline
\end{tabular}
\caption{Example counts as a function of the SNR proxy adapted for this work \S\ref{sec:obs_dataset}.  Classifier performance improved by $\sim$20\% by removing examples with DN$_{med}<$50, while for higher cuts the dataset is too sparse to train.  The distribution of DN$_{med}$ is plotted in Figure~\ref{fig:snr}.} 
\label{tbl:obs}
\end{table}

\section{This Work}
\label{sec:formulation}

We modify the initial convolutional layer of ResNet-152 to account for rectangular input data, exchanging kernel shape of 7x7 with 7x49 and stride from 2x2 to 2x12.  Kernel width showed no measurable effect on training or accuracy, suggesting a lack of importance in narrow spectral features.  Increasing stride in the spectral direction slightly reduces the capacity of the model while significantly reducing memory footprint, allowing larger input batches and increased training speed and stability.

For large datasets with 64 classes, a dropout rate of 10\% was selected empirically by varying from 0 (no dropout) to 20\% in steps of 2.5\%; accuracy slowly decreases above dropout rates of 15\% for this problem and architecture.  We hypothesize that the observed smooth decrease in accuracy is due to the decrease in effective model capacity with increasing dropout rate.  Epistemic uncertainty is measured by inferring 100 times for each validation inference.  This posterior mapping slows inference.  In space domain awareness we have the luxury of slow inference, but note that meaningfully descriptive uncertainty requires fewer forward passes.

We implement SWA, SWAG, multi-SWA and multi-SWAG and perform hyperparameter studies to settle on ideal model parameters before training ensembles.  We allow the scale of SWAG perturbations to vary, finding that values lower than 0.5 of \cite{Wilson2020BayesianGeneralization} are better suited to the dynamics of our problem.  These marginalization technqiues are applied to on sky and simulated nine class datasets.

\section{Datasets}
\label{sec:datasets}

\subsection{Simulated Data}
\label{sec:sim_dataset}

\begin{figure}[]
	\centering
	\includegraphics[width=0.48\textwidth]{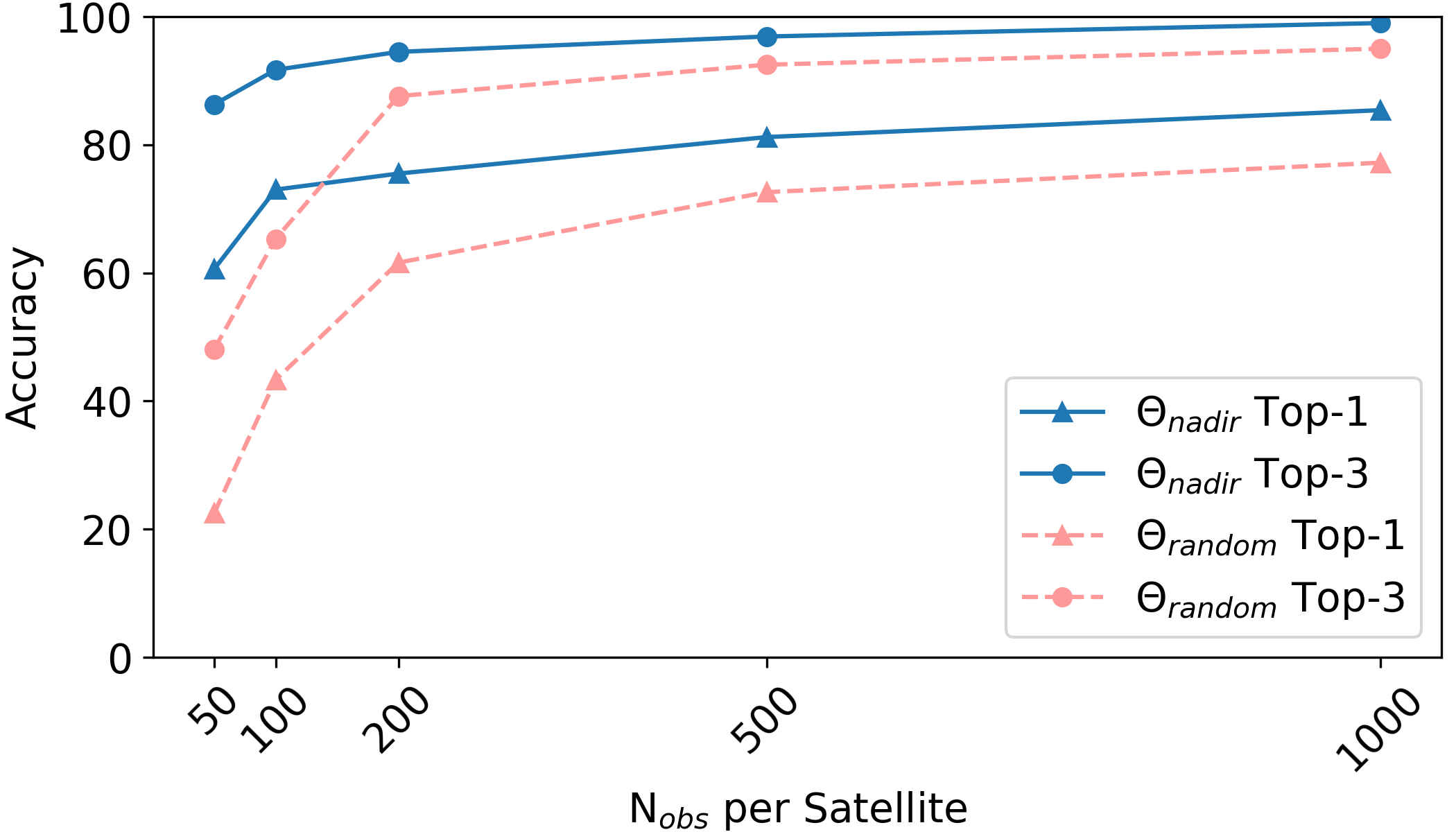}
	\caption{The effect of dataset size on classifier accuracy for a nominal, high sensitivity system trained on 64 satellite classes.  Blue solid lines represent constant nadir orientation ($\Theta_{nadir}$) while red dashed lines represent more difficult random orientations ($\Theta_{random}$).  Lines anchored by circles represent Top-3 accuracy, while triangles represent Top-1.  At few ($\sim$50) observations per target, accuracy between $\Theta_{nadir}$ and $\Theta_{random}$ are upwards of 40\%, a gap that shrinks to less than 10\% by 500 observations.}
	\label{fig:datasetsize}
\end{figure}

We simulate a spectrograph designed using commercial off the shelf optics components to enable realistic simulations of focal plane array (FPA) output and demonstrate the application of ResNets to affordable, standardized optical systems.  We utilize a proprietary radiometry code and adopt the Cerro Paranal Advanced Sky Model \cite{Jones2013AnParanal, Noll2012AnRange} to provide atmospheric transmission and emission based on parameters including precipitable water vapor (PWV), airmass, and observatory altitude.  Resulting images are 200x1340x1 in height, width, and channels (Figure $\ref{fig:examplespec}$).  Simulated images show a characteristic horizontal strip of exposed pixels; this strip is the result of an unresolved point spread function smeared along the horizontal image axis as a function of photon wavelength.  In this way, a single channel FPA resolves rich spectral$-$or color$-$information.  Our simulated instrument design captures a spectral energy distribution between 630 and 980 nanometers.  

We generate two families of datasets, one with all satellites maintaining nadir orientations ($\Theta_{nadir}$) in which the spacecraft are pointed directly towards Earth's surface.  This approximates normal operations for most space assets.  We also produce a more difficult dataset, in which every object randomly reorients between each exposure ($\Theta_{random}$).  This is an unrealistically challenging dataset, except for extremely high frame count datasets where the neural network is learning based on a complete sampling of the target orientation.  Reality falls somewhere in between$-$most space assets maintain a fixed orientation, but maneuvers and re-orientations do occur and are critical to identify through.  

Two observing modes, a nominal, dedicated instrument (64 classes at high signal to noise), and realistic early generation spectroscopic positive ID (nine classes at moderate signal to noise) are simulated, such that both current and future iterations of SpectraNet can be discussed.  Low signal to noise datasets are representative of the real data discussed in the following section.

\begin{table}[]
    \centering
    \begin{tabular}{| c | c | c | c | c |}
    \hline
       \multirow{3}{*}{N$_{obs}$ / RSO} & \multicolumn{4}{| c |}{ Classifier Accuracy } \\
        \cline{2-5} 
        &  \multicolumn{2}{| c |}{$\Theta_{nadir}$} &  \multicolumn{2}{| c |}{$\Theta_{random}$} \\
        & Top-1 & Top-3 & Top-1 & Top-3 \\
        \hline
        50 & 60.6 & 86.3 & 22.5 & 48.1 \\
        100 & 73.0 & 91.7 & 43.3 & 65.3 \\
        200 & 75.5 & 94.5 & 61.6 & 87.6 \\
        500 & 81.2 & 96.9 & 72.6 & 92.5 \\
        1000 & 85.4 & 99.0 & 77.2 & 95.0 \\
%        2000 & 86.9 & 99.3 & 81.3 & 96.7 \\
        \hline
    \end{tabular}
    \caption{Tabulated accuracies for the modified ResNet-152 64 class point estimate neural network described in \S\ref{sec:formulation}.  We use 80\% of data for training and hold out 10\% each for validation and testing. Baseline (random guessing) accuracy for a 64 class dataset is 1.56\%.}
    \label{tab:dataset_size}
\end{table}

\subsection{Observed Data}
\label{sec:obs_dataset}

We repeatedly observed eight RSOs over six nights between March and May of 2021 using the spectrograph simulated for \S\ref{sec:sim_dataset}.  Observing conditions ranged from photometric to scattered thin cloud cover.  We limited our observations to high altitudes and sunlit RSOs to maximize collected flux and minimize observing time. Exposure times vary between five seconds and five minutes.

A critical metric for applications of deep learning to scientific imagery is signal to noise ratio (SNR), a measure of the level of source signal compared to the observation noise.  For problems such as object detection \cite{Fletcher2019Feature-BasedNetworks}, decreasing SNR makes detections inherently more difficult as source signals are eventually overwhelmed by noise.  Spectroscopic SNR is generally measured as a function of wavelength after expert data calibration (e.g. \cite{Argyle2008AstronomicalAlgorithm}), but the \begin{it}implication\end{it} is the same -- for decreasing spectral SNR, individual spectral features become increasing difficult to detect.  In this work, we adopt a proxy for SNR, median count (DN$_{med}$), which provides a single signal measurement for a spectroscopic observation instead of a SNR spectrum,

We calculate DN$_{med}$ by taking the median count level across pixel rows in the dispersion direction, subtracting a polynomial fit to the background and bias levels, and summing the resulting counts, 

\begin{equation}
\textrm{DN}_{med} = \sum{DN_{spectral}} - DN_{bkg, spectral}
\end{equation}

DN$_{med}$ removes wavelength dependency on spectral SNR and is robust against outliers (hot pixels, cosmic rays), and spectral energy distribution.

\begin{figure}[]
	\centering
	\includegraphics[width=0.23\textwidth]{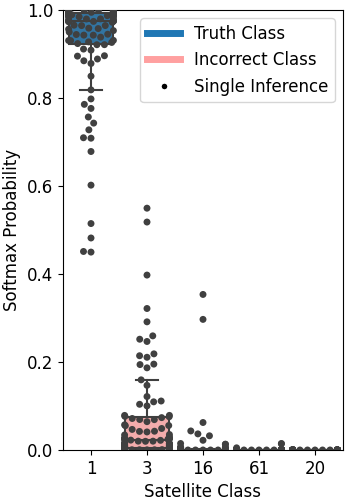}
	\includegraphics[width=0.23\textwidth]{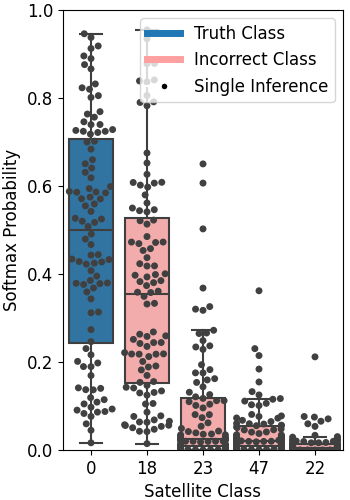}
	\caption{Bayesian inference showing the top-5 classifications for two validation set observations.  Black dots show single forward pass inferences repeated 100 times for the observation.  Shaded boxes denote the median softmax inference and surrounding quartiles; Dark blue denoting truth class, light red denoting incorrect classes.  \textbf{Left panel}: High confidence classification.  \textbf{Right panel}: a lower confidence classification.}
	\label{fig:bayes}
\end{figure}

For this work, DN$_{med}$ serves two additional roles.  First, the singular value per example illuminates classifier accuracy as a function of SNR, informing target limitations as a function of RSO size, distance and instrument collecting area.  Second, a single efficient proxy for SNR allows autonomous systems to quickly adjust exposure time to maximize the utility of collected spectra during unmanned observations.

The RSO observation statistics are tabulated in Table \ref{tbl:obs}, and include dataset sizes after a cut at DN$_{med} \geq$ 50 which significantly enhanced classifier performance, and $\geq$ 100, a limit which future experimentation will probe but for which we collected too few examples to exercise here.

We add 100 examples of an additional class in which the science instrument was exposed against no target.  Frames of this type are denoted as flats in the astronomy community and we adopt this nomenclature.  The addition of flats to this work bolsters dataset size and protects the classifier from improperly inferring a RSO class if observational conditions (misaligned instrument, cloud cover, instrument malfunction) result in an empty frame.  

\section{Experiments}
\label{sec:experiments}

We split this section into three experiments. The first (\S\ref{ssec:manyhigh}) explores simulated datasets with 64 classes and observations at high signal to noise.  This experiment represents an expectation of performance for a high quality, dedicated sensor powered by SpectraNet.  In this set of experiments, we rely on Bayesian marginalization by dropout due to large dataset sizes and the need for training efficiency.

In the second experiment, \S\ref{ssec:fewlow}, we describe results on simulated datasets of nine classes and moderate signal to noise observations, in line with our initial on sky dataset.  This section provides a theoretical baseline for our third experiment, described in \S\ref{sec:experiments_sky}, which contains results from actual spectroscopic data of artificial satellites at geostationary orbit.  For \S\ref{ssec:fewlow} and \S\ref{sec:experiments_sky} we adopt multiple variations of Stochastic Weight Averaging (SWA), which are well suited to boost performance on small datasets \cite{Izmailov2018AveragingGeneralization, Maddox2019ALearning, Wilson2020BayesianGeneralization}.  

We divide each section into two points of discussion, first, we discuss the impact of number of examples per class such that performance needs can be balanced against observing baselines for deployed SpectraNet systems.  Second, we discuss the performance of Bayesian marginalization against the datasets of each experiment.

\begin{table}[]
    \centering
    \begin{tabular}{| c | c | c | c |}
    \hline
    Threshold & \%$_{uncertain}$ & Top-1 Acc & Top-3 Acc  \\
    \hline
    \multicolumn{4}{ c }{200 obs 64 class $\Theta_{nadir}$} \\
    \hline
    0.4 & 16.2 & 75.0 & 96.1 \\
    0.6 & 28.6 & 80.4 & 97.2 \\
    0.8 & 43.2 & 87.1 & 98.1 \\
    \hline
    \multicolumn{4}{ c }{200 obs 64 class $\Theta_{random}$} \\
    \hline
    0.4 & 31.6 & 51.0 & 73.6 \\
    0.6 & 45.5 & 56.6 & 77.5 \\
    0.8 & 60.0 & 64.1 & 81.8 \\
    \hline
    \end{tabular}
    \caption{Tabulated accuracies for Bayesian models at varying thresholds of class probability.  Tuning up required softmax probability increases model accuracy by flagging inferences below the threshold value as uncertain classifications.  In high risk scenarios, knowing to ignore an inference is as critical as the prediction itself.}
    \label{tab:bayesian}
\end{table}

\subsection{Many Classes with High Signal to Noise}
\label{ssec:manyhigh}

In this experiment we simulate 64 class problems for random and nadir orientations.  The large number of classes allows reporting of Top-1 and Top-3 classifier accuracies.  While Top-1 results are impressive in their own regard in a field lacking performant positive identification techniques, the Top-3 results$-$in which the correct class is in a model's three most likely$-$are even more promising. In space traffic applications, two additional sources of information make top-N classification useful.  First, the nature of the top-N satellites is critical:  A Top-3 set of friendly assets is a different situation to unexpected or hostile assets.  In the latter case, followup observations or action can be taken.  Second, additional observational context is available for narrowing the top-N field.  For example, assets in a Top-3 set can be eliminated if known to be in orbits inconsistent with the observing geometry.

\begin{figure}[]
	\centering
	\includegraphics[width=0.48\textwidth]{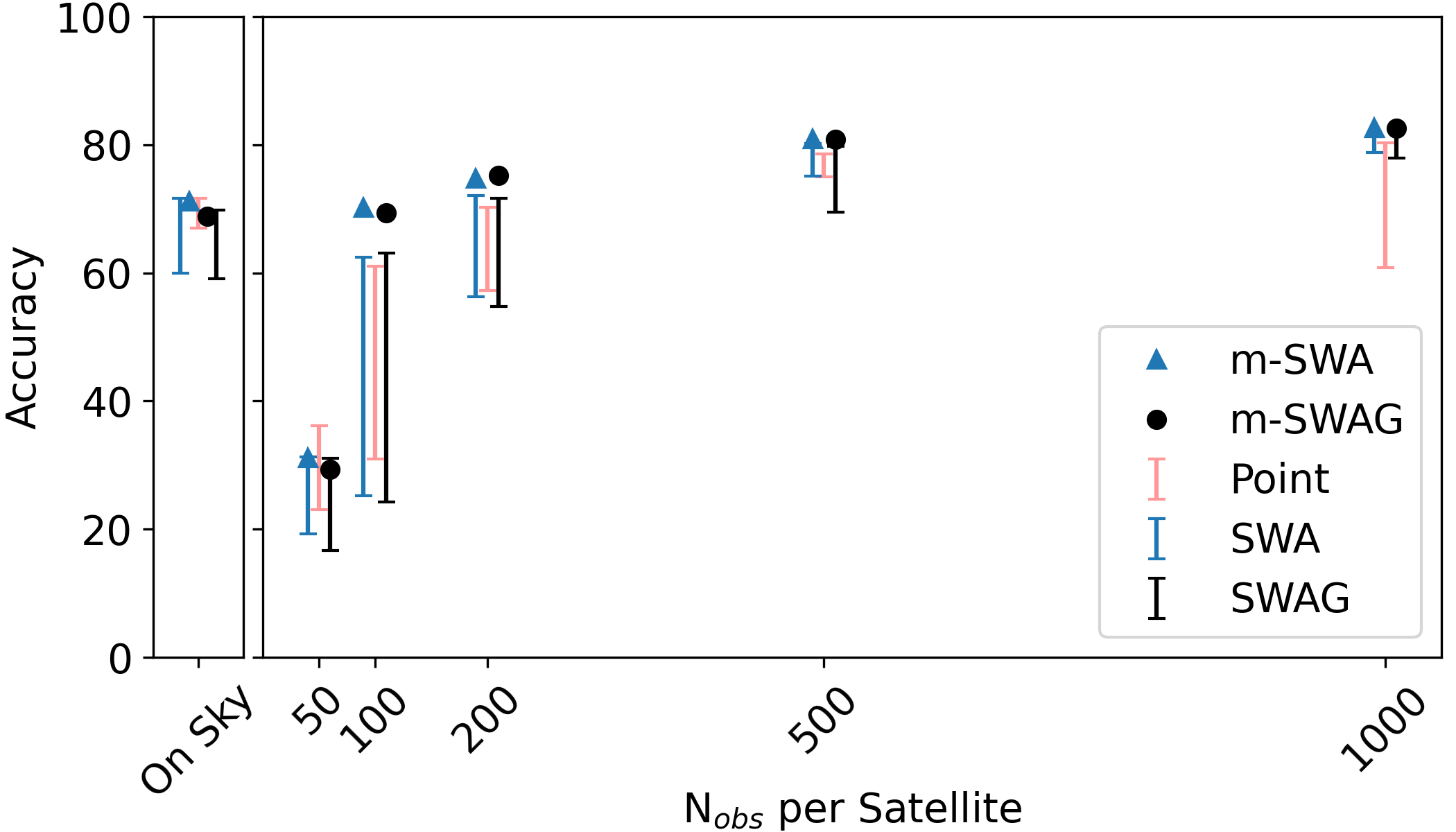}
	\caption{The effect of dataset size on classifier accuracy for a moderate sensitivity system trained on nine satellite classes on sky (\textbf{left panel}) and in simulation (\textbf{right panel}). 
	Red solid lines represent the accuracy performance bounds of 20 point estimate models.  Blue solid lines represent the bounds of the SWA models, and black solid lines bound the SWAG model accuracies.  Blue triangles mark the accuracy of Multi-SWA ensembles, and black dots show the accuracy of Multi-SWAG ensembles.}
	\label{fig:datasetsize_swa}
\end{figure}

\subsubsection{Number of Observations}
\label{ssec:manyhighnum}

Observing time on a ground based telescope is the limiting bottleneck for any data intensive effort.  Spectroscopic datasets grow slowly, at tens to hundreds of frames per night, even with a dedicated instrument under ideal atmospheric conditions.  This creates two considerations for this work: the accuracy of models trained on few examples, and the increase in accuracy as number of examples naturally grow as the technique is applied.  In Figure \ref{fig:datasetsize}, we visualize effects of dataset size on classifier accuracy (Table \ref{tab:dataset_size}).

The encouraging accuracy even at 50 observations per target is not unexpected.  While the diversity of class expression in classic computer vision problems is high, the spectroscopic images in this work directly encode reflection physics$-$in effect, the instrument itself represents a preprocessing step by which material and geometric features are orderly separated.  As dataset size grows, so does accuracy, especially in the case of random orientation where the network trains on an increasingly complete sampling of orientation distribution.  As a result this technique is quickly applicable and increases in efficacy with time.

\input{swa_table.tex}

\subsubsection{Bayesian Inference}
\label{ssec:manyhighbayes}

A forward pass through a point estimate classifier provides a single measurement of probability distribution across dataset classes which can be highly stochastic for small training datasets. By inferring repeatedly with dropout, we assemble a Monte Carlo ensemble of softmax distributions which map the uncertainty in classification given the model architecture and training dataset.

This statistically robust distribution provides inference uncertainty that is absent in point estimate networks.  Figure \ref{fig:bayes} visualizes the Top-5 softmax probability distributions for two inferences.  While both inferences are correct, a Bayesian model describes the uncertainty inherent so that actions based on inference are properly informed. 

In Table \ref{tab:bayesian} we demonstrate the tunable nature of Bayesian models.  By setting a threshold probability for inference, the model naturally specifies when inference is too uncertain to match requirements for informed action.  This boosts Top-1 and Top-3 accuracy by effectively ignoring suspect observations.  In practice, taking no action in high stakes environments can be preferred to mistake or accidental provocation.  The threshold value can be chosen per inference based on mission parameters.

For the challenging $\Theta_{random}$ dataset we find that our Bayesian implementation (Table \ref{tab:bayesian}) produces baseline accuracies $\sim$10\% below its point estimate counterpart (Table \ref{tab:dataset_size}).  Additional experimentation is underway to investigate this drop in performance, but we hypothesize that a dropout rate of 10 percent has lowered model capacity to below what is necessary for maximum classifier performance.  If this is the case, expanding model backbone capacity or investigating complex prior distributions over weights will correct this drop in performance.  In the $\Theta_{nadir}$ experiment, performance increases$-$as expected$-$by robustly classifying on median softmax distributions across ensembled classifiers.

\begin{figure}[]
	\centering
	\includegraphics[width=0.49\textwidth]{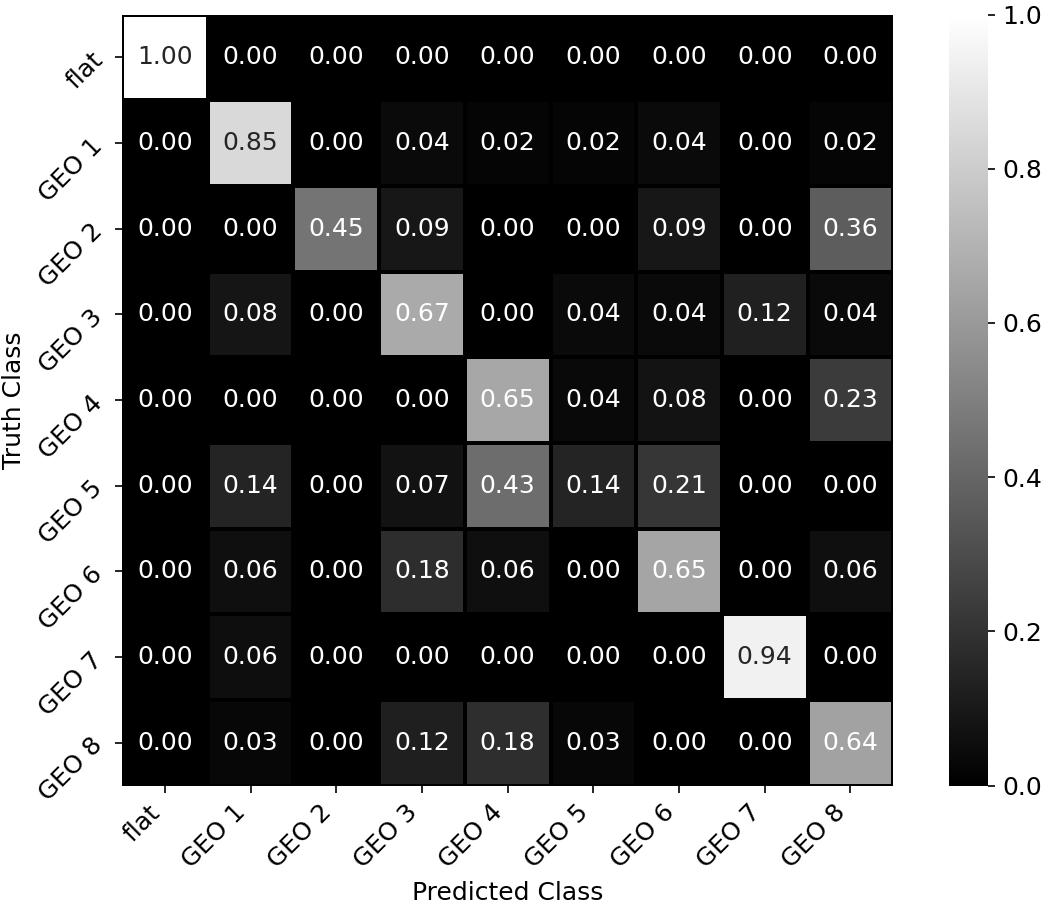}
	\caption{Confusion matrix for on sky validation examples.  This plot visualizes classification power in Table \ref{tbl:class_stats} and illuminates intra-class confusion.}
	\label{fig:conf_mat}
\end{figure}

\begin{table}
\centering\begin{tabular}{|l|rrr|r|}
\hline
  Class &  Precision &  Recall &    F1 &  Accuracy \\
\hline
  GEO 1 &      0.851 &   0.851 & 0.851 &         - \\
  GEO 2 &      1.000 &   0.455 & 0.625 &         - \\
  GEO 3 &      0.593 &   0.667 & 0.627 &         - \\
  GEO 4 &      0.548 &   0.654 & 0.596 &         - \\
  GEO 5 &      0.333 &   0.143 & 0.200 &         - \\
  GEO 6 &      0.550 &   0.647 & 0.595 &         - \\
  GEO 7 &      0.850 &   0.944 & 0.895 &         - \\
  GEO 8 &      0.618 &   0.636 & 0.627 &         - \\
   flat &      1.000 &   1.000 & 1.000 &         - \\
\hline
Overall &          - &       - &     - &     0.716 \\
\hline
\end{tabular}
\caption{Classifier Performance Statistics.  Overall classifier accuracy of 71.6\% agrees with theoretical expectations in \S\ref{ssec:fewlow}, as training examples vary from 54 to 235 (Table \ref{tbl:obs})}
\label{tbl:class_stats}
\end{table}

\subsection{Few Classes at Moderate Signal}
\label{ssec:fewlow}

Here we simulate a nine class dataset embracing more moderate values of signal to noise than in the previous section.  This dataset uniformly samples DN$_{med}$ values between 50 and 1000 to closely match the on sky dataset (see Figure \ref{fig:snr}) while providing more complete sampling of the space.   Due to the small number of classes, we report only Top-1 classifier performances.

We train 20 identical models for each ensembling experiment, selecting model parameters (learning rate, kernel regularization) from a hyperparameter study performed before each ensembling training run.  Individual models are trained from scratch for 300 epochs, with SWA mean and SWAG covariances tracked over the last 20 epochs.  Results for these runs are tabulated in Table \ref{tab:swag} and presented, for the $\Theta_{nadir}$ case, in Figure \ref{fig:datasetsize_swa}.  For point estimate, SWA, and SWAG values, which are calculated per-model, we tabulate the highest performing iteration and plot the extent of the model performances.  Multi-SWA and Multi-SWAG are ensembled techniques$-$the reported value is a singular measured performance.

\begin{figure*}[]
	\centering
	\includegraphics[width=0.49\textwidth]{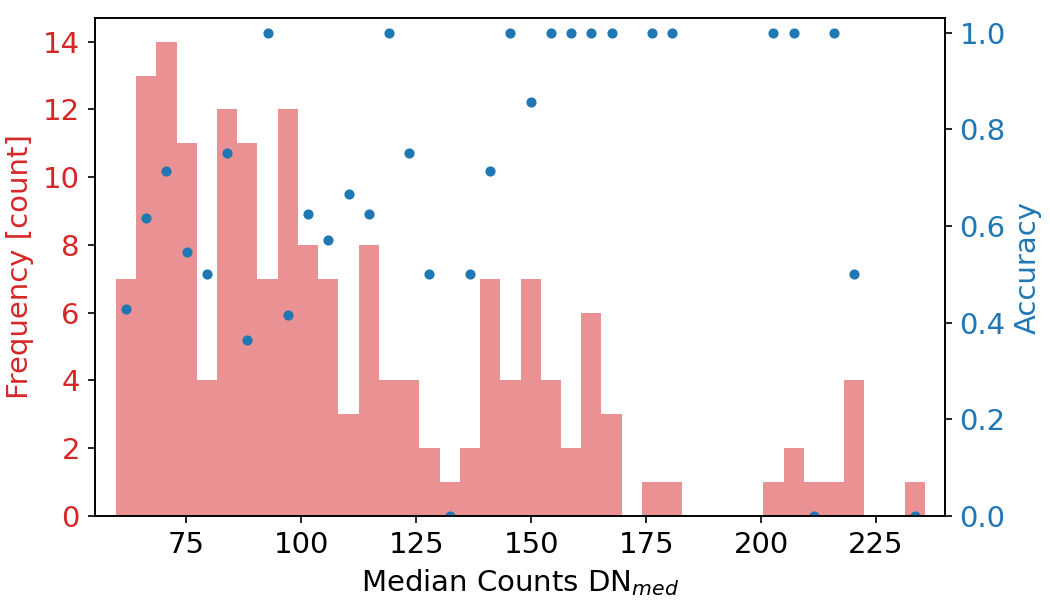}	\includegraphics[width=0.49\textwidth]{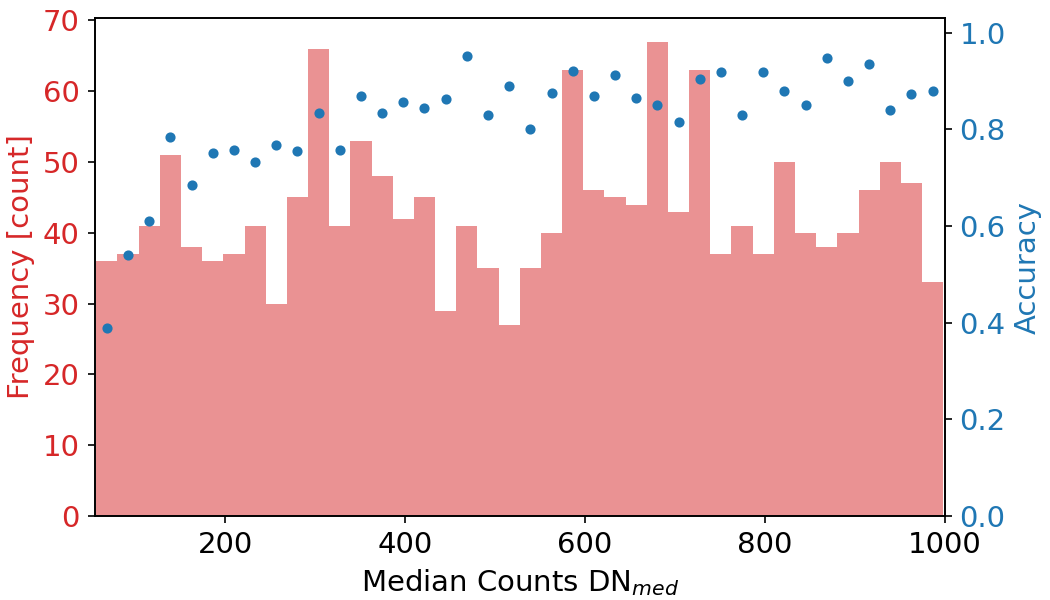}
	\caption{Model validation accuracy as a function of DN$_{med}$ in on sky data (left panel) and simulated data at 500 examples per class (right panel).  Even with the limited observed dataset size in this work, a trend between classifier accuracy and DN$_{med}$ is becoming apparent.  This drop off in performance with decreased DN$_{med}$ agrees with simulation and elucidates a pareto front optimization between sensor efficiency, target brightness, and observing time critical to designing practical SpectraNet powered systems.}
	\label{fig:acc_v_snr}
\end{figure*}
\subsubsection{Number of Observations}
\label{ssec:fewlownum}

As in \S\ref{ssec:manyhigh}, classifier performance improves dramatically with increased observations, which is apparent in both Table \ref{tab:swag} and Figure \ref{fig:datasetsize_swa}.  The performance loss between our many class work and this experiment is expected, but confounded between larger overall dataset in the prior, and lower signal in the latter.  For each SpectraNet scenario, inter-class spectral signature differences$-$the difficulty of the particular problem based on satellite similarity$-$also plays an unexplored role.

Still, these experiments show that accuracies over 80\% can be expected for deployed SpectraNet systems even without ideal instruments or observing conditions.

\subsubsection{Bayesian Inference}
\label{ssec:fewlowbayes}

Once observations per class reach 100, multi-basin Bayesian marginalization techniques outperform point estimate and single-basin formulations.  This result is in line with previous work in the field \cite{Izmailov2018AveragingGeneralization, Maddox2019ALearning, Wilson2020BayesianGeneralization}.  We hypothesize that while multiple basins of attraction can yield similar overall classifier accuracy, unique basins provide unique perspectives on the problem.  Marginalization over these perspectives both improves classifier accuracy and decreases the miscalibrated sharp overconfidence of point estimate models.  In fact, we commonly find that point estimate models are overconfident while ensembled methods are underconfident.  In both scenarios, tempering lowers ECE to reasonable values.

\subsection{On Sky Data}
\label{sec:experiments_sky}

We run the same experiment discussed in \ref{ssec:fewlow} on 90\% of examples from our on sky dataset with DN$_{med} >$ 50, holding 10\% out to measure the validation statistics reported here.  In Table \ref{tbl:class_stats} and Figure \ref{fig:conf_mat} we demonstrate classifier accuracy statistics and visualize the classifier confusion matrix, respectively.  SpectraNet trains on an average of $\sim$50 to 230 observations per class to an overall accuracy of 71.6\%.  We note that this performance is in strong agreement with our simulated experiment on number of observations in \S\ref{ssec:fewlow}, where the 100 and 200 example problems reach accuracies of 70.2\% and 75.2\%, respectively.  Single-basin techniques outperform multi-basin in this problem.  We assert that this is caused by the low ($<$100) example counts for four of the classes.

Classes with fewest examples (GEOs 2 and 5) perform poorly, while the third (GEO 4) is heavily weighted towards lower DN$_{med}$.  These classes are most often confused with the highest population classes, GEO 1 and 8, as seen in Figure \ref{fig:conf_mat}.  We have shown in \S\ref{ssec:fewlow} that performance degrades significantly by $\sim$50 observations per class.

In Figure \ref{fig:acc_v_snr} we measure classifier validation accuracy across bins of DN$_{med}$ and argue that increasing accuracy is correlated with increasing signal to noise.  Our simulted results clearly demonstrate this hypothesis.

\section{Conclusions}
\label{sec:conclusion}

% In this work we are confronted with a problem space in which data collection is difficult, and in which inference uncertainty is important.  However, training efficiency is also critical in space domain awareness, where computation can be limited and models need to be trained against rapidly evolving on-orbit conditions. 

We demonstrate a learned spectroscopic positive identification framework for artificial satellites by modeling a high yield, low cost sensor and training convolutional neural networks on raw simulated output.  By avoiding physics-based priors and data calibration or reduction, we demonstrate the practicality of this approach.  

In this work we demonstrate that Top-1 accuracies of 80-90\% and Top-3 of over 99\% are achievable for many-satellite identification models.  We show that models are performant with $\sim$50 observations per class, and that performance increases steadily with continued observations.

We demonstrate the applicability of Bayesian marginalization, and describe the importance of model and data uncertainty in a field with significant action space risk.

Finally, we collect a spectroscopic dataset of artificial satellites at geostationary orbit and show that models trained using on sky data are performant at accuracies of $\sim$72\% after 100 observations per target, in encouraging agreement with expectations from simulation.

The impact of machine intelligence solutions to the problems of space domain awareness relying on scientific imagery will continue to grow with the proliferation of large astronomical survey missions and increasing density of orbital assets.  We present this work on spatially unresolved positive identification as the baseline on which we hope this community will build.

{\small
\bibliographystyle{ieee_fullname}
\bibliography{references}
% \bibliography{egbib}
}

\end{document}

%% file: swa_table.tex
\begin{table}[]
\centering
\begin{tabular}{| l | c | c | c | c | c | c |}
\hline
\multirow{2}{*}{Eval} &  \multicolumn{3}{| c |}{$\Theta_{nadir}$} & \multicolumn{3}{| c |}{$\Theta_{random}$} \\
    & Acc & ECE & T & Acc & ECE & T \\
\hline
\multicolumn{7}{ c }{On Sky Observations $\sim$100-300 examples / class} \\
\hline
Point & \textbf{71.6} & 0.10 & 1.25 & - & - & - \\
SWA & \textbf{71.6} & 0.08 & 1.50 & - & - & - \\
m-SWA & 71.2 & 0.07 & 1.10 & - & - & - \\
SWAG & 69.8 & 0.10 & 0.90 & - & - & - \\
m-SWAG & 68.8 & 0.07 & 0.55 & - & - & - \\
\hline
\multicolumn{7}{ c }{Simulated 50 examples/class} \\
\hline
Point & \textbf{36.3} & 0.06 & 1.60 & \textbf{25.9} & 0.07 & 6.60 \\
SWA & 31.3 & 0.04 & 7.85 & 24.0 & 0.05 & 8.00 \\
m-SWA & 31.2 & 0.04 & 1.15 & 20.9 & 0.03 & 1.05 \\
SWAG & 31.1 & 0.04 & 5.60 & 23.4 & 0.04 & 2.30 \\
m-SWAG & 29.3 & 0.04 & 1.05 & 20.9 & 0.03 & 0.90 \\
\hline
\multicolumn{7}{ c }{Simulated 100 examples/class} \\
\hline
Point & 61.0 & 0.03 & 2.20 & 36.4 & 0.04 & 3.35 \\
SWA & 62.5 & 0.03 & 3.00 & 33.4 & 0.05 & 6.00 \\
m-SWA & \textbf{70.2} & 0.03 & 0.55 & 37.6 & 0.07 & 0.45 \\
SWAG & 62.3 & 0.05 & 1.05 & 32.6 & 0.06 & 4.10 \\
m-SWAG & 69.4 & 0.04 & 0.60 & \textbf{37.8} & 0.06 & 0.40 \\
\hline
\multicolumn{7}{ c }{Simulated 200 examples/class} \\
\hline
Point & 70.2 & 0.04 & 2.15 & 49.4 & 0.05 & 2.05 \\
SWA & 72.1 & 0.03 & 2.00 & 45.8 & 0.06 & 3.80 \\
m-SWA & 74.8 & 0.03 & 0.80 & 51.6 & 0.06 & 0.25 \\
SWAG & 71.6 & 0.04 & 0.50 & 46.2 & 0.06 & 3.45 \\
m-SWAG & \textbf{75.2} & 0.03 & 0.65 & \textbf{52.1} & 0.05 & 0.20 \\
\hline
\multicolumn{7}{ c }{Simulated 500 examples/class} \\
\hline
Point & 78.6 & 0.02 & 1.10 & 64.4 & 0.02 & 1.35 \\
SWA & 80.2 & 0.02 & 0.95 & 65.9 & 0.03 & 1.25 \\
m-SWA & \textbf{80.9} & 0.03 & 0.50 & \textbf{67.8} & 0.04 & 0.20 \\
SWAG & 79.8 & 0.02 & 0.75 &  66.1 & 0.03 & 0.65 \\
m-SWAG & 80.8 & 0.01 & 0.50 & 67.4 & 0.03 & 0.15 \\
\hline
\multicolumn{7}{ c }{Simulated 1000 examples/class} \\
\hline
Point & 80.3 & 0.03 & 1.00 & 69.7 & 0.03 & 0.95 \\
SWA & 81.3 & 0.03 & 1.00 & 70.3 & 0.03 & 0.80 \\
m-SWA & \textbf{82.7} & 0.03 & 0.50 & \textbf{71.7} & 0.04 & 0.15 \\
SWAG & 82.1 & 0.03 & 0.50 & 70.7 & 0.03 & 0.60 \\
m-SWAG & 82.6 & 0.02 & 0.50 & 71.2 & 0.04 & 0.15 \\
    \hline
\end{tabular} 
\caption{Tabulated accuracies for classifiers trained on raw spectroscopic frames of nine moderate signal resident space objects.  Each table panel is described by it's header: whether or not the data is simulated or on sky, and how many examples per class the models were trained on.  In each panel, the best performing technique(s) are noted in bold face.} 
\label{tab:swag} 
\end{table}